\documentclass{article} 
\usepackage{iclr2020_conference,times}

\usepackage{amsmath,amsfonts,bm}

\def\eqref#1{equation~\ref{#1}}

\def\1{\bm{1}}

\DeclareMathAlphabet{\mathsfit}{\encodingdefault}{\sfdefault}{m}{sl}
\SetMathAlphabet{\mathsfit}{bold}{\encodingdefault}{\sfdefault}{bx}{n}

\usepackage{pgfplots}
\usepackage{hyperref}
\usepackage{url}
\usepackage[english]{babel}
\usepackage{multirow}
\usepackage{subfig}
\usepackage[colorinlistoftodos]{todonotes}
\usepackage{booktabs}

\title{GDP: Generalized Device Placement for Dataflow Graphs}

\author{Yanqi Zhou, Sudip Roy, Amirali Abdolrashidi\thanks{Work done during internship at Google}, Daniel Wong\thanks{Work done during internship at Google}, Peter C. Ma, Qiumin Xu \AND Ming Zhong, Hanxiao Liu, Anna Goldie, Azalia Mirhoseini, James Laudon  \\
\\
Google Brain\\
\texttt{\{yanqiz,sudipr,abdolrashdi,wonglkd,pcma,qiuminxu\}@google.com} \\
\texttt{\{mingzhong,hanxiaol,agoldie,azalia,jlaudon\}@google.com} \\
}

\iclrfinalcopy 
\begin{document}

\maketitle

\begin{abstract}

Runtime and scalability of
large neural networks can be significantly affected by the placement of operations in their dataflow graphs on suitable devices. With increasingly complex neural network architectures and heterogeneous device characteristics, finding a reasonable placement
is extremely challenging even for domain experts. Most existing automated device placement approaches are impractical due to the significant amount of compute required
and their inability to generalize to new, previously held-out graphs. To address both limitations, we propose an efficient end-to-end method based on
a scalable sequential attention mechanism over a graph neural network that is transferable to new graphs.
On a diverse set of representative deep learning models, including Inception-v3, AmoebaNet, Transformer-XL, and WaveNet,
our method on average achieves 16\% improvement over human experts and 9.2\% improvement over
the prior art
with 15$\times$ faster convergence.
To further reduce the computation cost,
we pre-train the policy network on a set of dataflow graphs and use a superposition network to fine-tune it on each individual graph,
achieving state-of-the-art performance on large hold-out graphs with over 50k nodes, such as an 8-layer GNMT.
\end{abstract}

\section{Introduction}
\label{intro}

Neural networks have demonstrated remarkable scalability--improved performance can usually be achieved by training a larger model on a larger dataset \citep{hestness2017deep,shazeer2017outrageously,jozefowicz2016exploring,mahajan2018exploring,radford2019language}.  
Training such large models efficiently while meeting device constraints, like memory limitations, necessitate partitioning of the underlying dataflow graphs for the models across multiple devices. However, devising a good partitioning and placement of the dataflow graphs  requires deep understanding of the model architecture, optimizations performed by domain-specific compilers, as well as the device characteristics, and is therefore extremely hard even for experts.

ML practitioners often rely on their understanding of model architecture to determine a reasonable partitioning and placement for graphs. However, relying solely on the model architecture while ignoring the effect of the partitioning on subsequent compiler optimizations like op-fusion can lead to sub-optimal placements and consequently under-utilization of available devices. The goal of automated device placement is to find the optimal assignment of operations to devices such that the end-to-end execution time for a single step is minimized and all device constraints like memory limitations are satisfied. Since this objective function is non-differentiable, prior approaches~\citep{Mirhoseini2017ICML,Mirhoseini2018ICLR,pmlr-v80-gao18a} have explored solutions based on reinforcement learning (RL). However, these RL policies are usually not transferable and require training a new policy from scratch for each individual graph. This makes such approaches impractical due to the significant amount of compute required for the policy search itself, at times offsetting gains made by the reduced step time.

In this paper, we propose an end-to-end deep RL method for device placement where the learned policy is generalizable to new graphs. Specifically, the policy network consists of a graph-embedding network that encodes operation features and dependencies into a trainable graph representation, followed by a scalable sequence-to-sequence placement network based on an improved Transformer \citep{NIPS2017_7181, Zihang2019ACL}.
The placement network transforms the graph representations into a placement decision with soft attention, removing hard constraints such as hierarchical grouping of operations \citep{Mirhoseini2018ICLR} or co-location heuristics (to reduce the placement complexity) \citep{Mirhoseini2017ICML}.
Both of our graph-embedding network and placement network can be jointly trained in an end-to-end fashion using a supervised reward, without the need to manipulate the loss functions at multiple levels. We empirically show that the network learns flexible placement policies at a per-node granularity and can scale to problems over 50,000 nodes.

To generalize to arbitrary and held-out graphs, our policy is trained jointly over a set of dataflow graphs (instead of one at a time) and then fine-tuned on each graph individually. By transferring the learned graph embeddings and placement policies, we are able to achieve faster convergence and thus use less resources to obtain high-quality placements.
We also use super-positioning, i.e., a feature conditioning mechanism based on the input graph embeddings, to effectively orchestrate the optimization dynamics of graphs with drastically different sizes in the same batch.

Our contributions can be summarized as follows:
\begin{enumerate}
    \item An end-to-end device placement network that can generalize to arbitrary and held-out graphs. This is enabled by jointly learning a transferable graph neural network along with the placement network.
    \item A scalable placement network with an efficient recurrent attention mechanism, which eliminates the need for an explicit grouping stage before placement.
    The proposed end-to-end network provides 15$\times$ faster convergence as compared to the hierarchical LSTM model used in earlier works~\citep{Mirhoseini2017ICML,Mirhoseini2018ICLR}. 
    \item A new batch pre-training and fine-tuning strategy based on network superposition,
    which leads to improved transferability,
    better placements especially for larger graphs, and 10$\times$ reduction
    in policy search time as compared to training individual graphs from scratch.
    \item Superior performance over a wide set of workloads, including InceptionV3~\citep{Inception2015}, AmoebaNet~\citep{amoeba2018}, RNNs, GNMT~\citep{GNMT2016}, Transformer-XL \citep{Zihang2019ACL},  WaveNet~\citep{wavenet2016}, and more.
\end{enumerate}

\section{Related Work}
\label{related}

\paragraph{Device Placement}
Reinforcement learning has been used for device placement of a given dataflow graph \citep{Mirhoseini2017ICML} and demonstrated run time reduction over human crafted placement and conventional heuristics. For improved scalability, a hierarchical device placement strategy (HDP) \citep{Mirhoseini2018ICLR} has been proposed that clusters operations into groups before placing the operation groups onto devices. 
Spotlight~\citep{pmlr-v80-gao18a} applies proximal policy optimization and cross-entropy minimization to lower training overhead.
Both HDP and Spotlight rely on LSTM controllers that are difficult to train
and struggle to capture very long-term dependencies
over large graphs.
In addition, both methods are restricted to process only a single graph at a time, and cannot generalize to arbitrary and held-out graphs.
Placeto~\citep{Addanki2019Placeto} represents the first attempt to generalize device placement using a graph embedding network. But like HDP, Placeto also relies on hierarchical grouping and only generates placement for one node at each time step.
Our approach (GDP) leverages a recurrent attention mechanism and generates the whole graph placement at once. This significantly reduces the training time for the controller. We also demonstrate the generalization ability of GDP over a wider set of important workloads.

\paragraph{Parallelization Strategy}
Mesh-TensorFlow is a language that provides a general class of distributed tensor computations. While data-parallelism can be viewed as splitting tensors and operations along the ``batch'' dimension, in Mesh-TensorFlow the user can specify
any tensor-dimensions to be split across any dimensions of a multi-dimensional mesh of processors. FlexFlow \citep{Jia2018sysml} introduces SOAP, a more comprehensive search space of parallelization strategies for DNNs which allows parallelization of a DNN in the Sample, Operator, Attribute, and Parameter dimensions. It uses guided randomized search of the SOAP space to find a  parallelization strategy for a specific parallel machine. GPipe~\citep{Gpipe2018cvpr} proposed pipeline parallelism, by partitioning a model across different accelerators and automatically splitting a mini-batch of training examples into smaller micro-batches. By pipelining the execution across micro-batches, accelerators can operate in parallel. Our GDP focuses on a general deep RL method for automating device placement on arbitrary graphs, and is therefore orthogonal to existing parallelization strategies.

\paragraph{Compiler Optimization}
REGAL \citep{Aditya2019KDD} uses deep RL to optimize the execution cost of computation graphs in a static compiler. The method leverages the policy's ability to transfer to new graphs to improve the quality of the genetic algorithm for the same objective budget. However, REGAL only targets peak memory minimization while GDP focuses on graph run time and scalability while also meeting the peak memory constraints of the devices. Specifically, we generalize graph partitioning and placement into a single end-to-end problem, with and without simulation, which can handle graphs with over 50,000 nodes.

\section{End-to-End Placement Policy}
\label{sec:method}
Given a dataflow graph $G(V,E)$ where $V$ represents atomic computational operations (ops) and $E$ represents the data dependency, our goal is to learn a policy $\pi: \mathcal{G} \mapsto \mathcal{D}$ that assigns a placement $D \in \mathcal{D}$
for all the ops in the given graph $G \in \mathcal{G}$, to maximize the reward $r_{G, D}$ defined based on the run time.
In this work, we represent policy $\pi_\theta$ as a neural network parameterized by $\theta$.

Unlike prior works that focus on a single graph only,
the RL objective in GDP is defined to simultaneously reduce the expected runtime of the placements
over a set of $N$ dataflow graphs:

\begin{align}
  \label{eqn:cost}
  J(\theta) = \mathbb{E}_{G \sim \mathcal{G}, D \sim \pi_\theta(G)}[r_{G,D}]
  \approx
  \frac{1}{N}\sum_{G} \mathbb{E}_{D \sim \pi_\theta(G)} [r_{G, D}]
\end{align}

In the following, we refer to the case when $N=1$ as individual training and the case when $N>1$ as \emph{batch training}.
We optimize the objective above using Proximal Policy Optimization (PPO)~\citep{PPO2017} for improved sample efficiency.

\label{subsec:end-to-end}
\begin{figure}[t]
\begin{center}
 \includegraphics[width=0.9\textwidth,trim=0cm 5cm 0cm 5cm, clip]{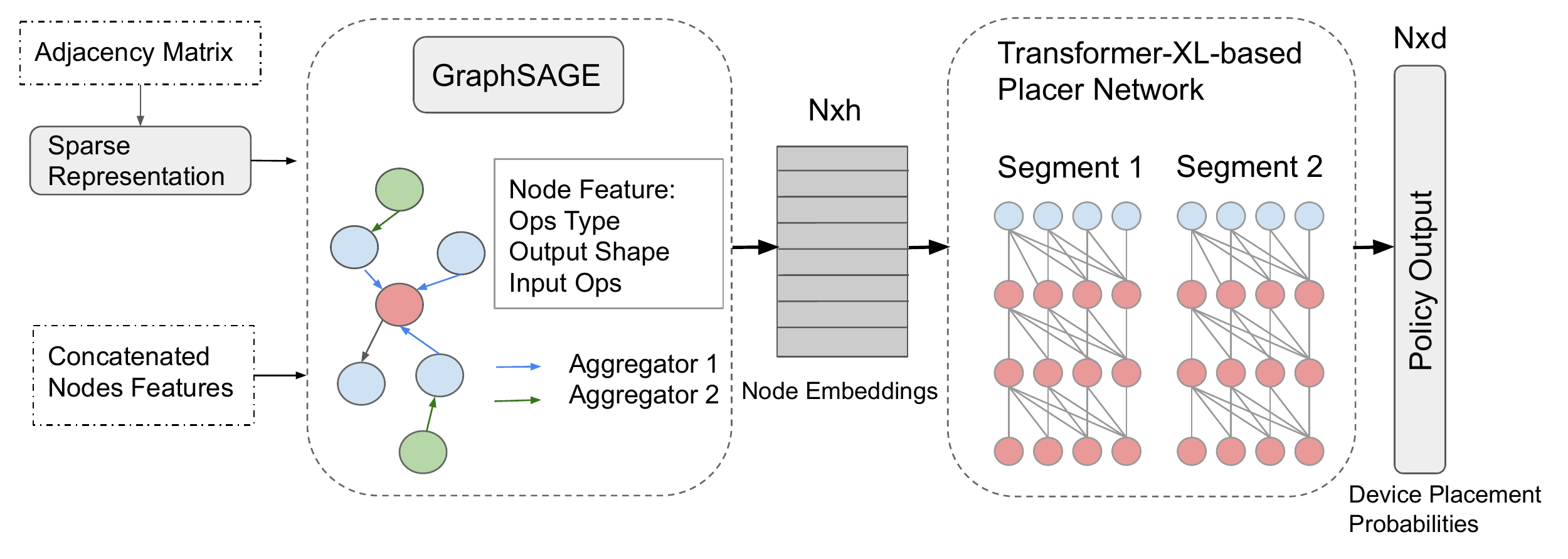}
\end{center}
\caption{Overview of GDP: An end-to-end placement network that combines graph embedding and sequential attention. $N$: Number of Nodes, $h$: Hidden Size, $d$: Number of Devices.}
\label{fig:end-to-end}
\end{figure}

Our proposed policy network $\pi_\theta$ consists a graph embedding network that learns the graphical representation of any dataflow graph,
and a placement network that learns a placement strategy over the given graph embeddings.
The two components are jointly trained in an end-to-end fashion.
Note the architecture is designed to be invariant over the underlying graph topology, enabling us to apply the same learned policy
to a wide set of input graphs with different structures.

Figure~\ref{fig:end-to-end} shows an overview of the proposed end-to-end device placement network.
Next, we will discuss on each of the components in details.


\subsection{Graph Embedding Network}
We leverage graph neural networks (GNNs) \citep{William2017NIPS,Keyulu2019ICLR,GCPN2018nips} to capture the topological information encoded in the dataflow graph. GNNs have gained increasing popularity in various domains for their ability in learning efficient representations over graph-structured and relational data. However, most graph embedding frameworks are inherently transductive and can only generate embeddings for a given fixed graph. These transductive methods do not efficiently extrapolate to handle unseen nodes (e.g., in evolving graphs), and cannot learn to generalize to unseen graphs. GraphSAGE~\citep{William2017NIPS} is an inductive framework that leverages node attribute information to efficiently generate representations on previously unseen data.
While our proposed framework is generic, we adopt the feature aggregation scheme proposed in GraphSAGE to model the dependencies between the operations and build a general, end-to-end device placement method for a wide set of dataflow graphs.


In GDP,
nodes and edges in the dataflow graph are represented as the concatenation of their meta features (e.g., operation type, output shape, adjacent node ids) and are further encoded by the graph embedding network into a trainable representation.
The graph embedding process consists of multiple iterations, and the computation procedure for the $l$-th iteration can be outlined as follows:

First, each node $v \in V$ aggregates the feature representations of its neighbors, $\{h_{u}^{(l)}, \forall u\in \mathcal{N}(v)\}$, into a single vector $h_{\mathcal{N}(v)}^{(l)}$. This aggregation
outcome is a function of all previously generated representations,
including the initial representations defined based on the input node features.
In this work, we use the following aggregation function with max pooling:
\begin{align}
h_{\mathcal{N}(v)}^{(l)} =
\max(\sigma(W^{(l)}h_{u}^{(l)} + b^{(l)}), \forall u \in \mathcal{N}(v))
\end{align}
where $(W^{(l)}$, $b^{(l)})$ define an affine transform and $\sigma$ stands for the sigmoid activation function.
We then concatenate the node's current representation, $h_{v}^{(l)}$, with the aggregated neighborhood vector, $h_{\mathcal{N}(v)}^{(l)}$, and feed this concatenated vector through a fully connected layer $f^{(l+1)}$
\begin{align}
    h_{v}^{(l+1)}=f^{(l+1)}(\mathrm{concat}(h_{v}^{(l)}, h_{\mathcal{N}(v)}^{(l)}))
\end{align}
Different from GraphSAGE, parameters in our graph embedding network are trained jointly with a placement network via stochastic gradient descent with PPO, in a \emph{supervised} fashion, as described in Section~\ref{sec:method}. That is, we replace the unsupervised loss with our task-specific objective.  

\subsection{Placement Network}
Designing a scalable placement network that can generalize to graphs with thousands of nodes is challenging, as the conventional GNMT models proposed for language tasks usually target a shorter sequence length. Hierarchical placement~\citep{Mirhoseini2018ICLR} has been proposed to address this issue,however, the proposed grouper network comes with limited flexibility and generality. For example, the grouper network leverages an aggregated feature representation by averaging feature vectors for nodes within the same group. The non-differentiable grouping procedure prevents training the graph-embedding and placement networks end-to-end. 

To remove the two-stage hierarchical workflow in HDP for improved scalability, we propose to use a
Transformer-based attentive network to generate operation placements in an end-to-end fashion. As the graph embedding already contains spatial (topological) information for each node, we remove the positional embedding in the original transformer to prevent the model from overfitting node identifications.
To capture long-term dependencies efficiently among a large set of nodes,
we adopt segment-level recurrence introduced in Transformer-XL ~\citep{Zihang2019ACL,dai2019improving},
where hidden states computed for the previous set of nodes are cached (with gradient flows disabled) and reused as an extended context during the training of the next segment.
Besides achieving extra long context, we empirically find the segment-level recurrent attention much faster than a conventional LSTM-based GNMT model.
In our experimental evaluation, we compare both the performance and speed up of our placement network with that of the LSTM-based hierarchical device placement. 

\subsection{Batch Training with Parameter Superposition}
Since the parameterization for the architecture of the end-to-end policy is designed to be invariant over input graphs with different topologies, the same placement policy can be shared across a wide set of workloads. We therefore propose a batch training strategy, and further enhance the aforementioned architecture
to handle such generalization across graphs.

Na\"ive batch training is challenging in our context as different dataflow graphs contain different number of operations connected in different topologies.
In addition, unlike previous device placement methods,
GDP aims to handle graphs from potentially different application domains (e.g. computer vision, language, and speech), where the number of operations can range from a few thousand to one million. These graphs have drastically different network architecture, in terms of computational operations, data shape, and network topology. As an example, recurrent networks have completely different operation types and connections compared to multi-branch convolutional networks that are widely used in computer vision. It would be highly desirable to train a single shared network that maximizes information sharing across these heterogeneous tasks, without hurting the performance on each of them due to their distinct learning dynamics.

To this end, we propose a feature conditioning mechanism similar to \emph{parameter superposition}~\citep{superposition2019Brian}.
The idea is to train one shared policy,
but condition its parameters based on the input features to mitigate the potentially undesirable interference among different input graphs.
Since dense layers (affine transforms followed by nonlinearity) serve as the fundamental building blocks in all of our network components, we introduce an additional conditioning layer to enable superposition in all dense layers the placement network:
\begin{align}
    x^{(l+1)}=g^{(l)}(c(x^{(0)})\odot x^{(l)})
\end{align}
where $g^{(l)}$ stands for a dense layer in our policy network, $c$ stands for the feature conditioning layer, and $x^{(0)}$ denotes the feature representation of the input graph generated by the graph-embedding network.
The feature conditioning layer is implemented with minimum overhead by adding an additional transformer layer to our placement network.


\section{Experiment}
\label{sec:experiment}
\subsection{Experiment Setup}
In this section, we evaluate our training strategy on widely used machine learning models in computer vision, natural language processing, and speech domains. We compare our approach to human expert placement, TensorFlow METIS placement, and hierarchical device placement (HDP)~\citep{Mirhoseini2018ICLR}.
Our experiments are run on machines with one Intel Broadwell CPU and up to eight Nvidia P100 GPUs. Note that the prior work~\citep{Mirhoseini2017ICML,Mirhoseini2018ICLR,pmlr-v80-gao18a} were evaluated on different GPU devices, preventing direct comparison of results. Therefore, we re-evaluate HDP on our own system environment and report those numbers. 

The performance of a placement is evaluated by the resulted training step time (run time) of the neural network. We use the negative square root of the run time as the reward. We use the average reward of all the previous trials as a bias term. The advantage value is computed by subtracting the reward by the average reward. During the search, we apply a large negative reward (-10) for invalid placements (e.g. a violation of co-location constraint, out of memory, etc.).

\subsection{Performance on Individual Graphs}
\label{subsec:ind_perf}
We evaluate GDP by training the model separately on six important graphs, including RNN Language Modeling, GNMT~\citep{seq2seq2014}, Transformer-XL, Inception, AmoebaNet, and WaveNet. We name this approach \textbf{GDP-one}. For all the tasks, GDP-one consistently outperforms human expert placement, TensorFlow METIS~\citep{Karypis:1998:FHQ:305219.305248} placement, and HDP. For extremely large graphs, GDP-one is only 6\% worse on 8-layer NMT (over 60k nodes), compared to human placement, but is 6.8\% better than HDP. Overall, GDP-one achieves on average more than 16\% run time reduction across the evaluated 12 graphs, compared to human expert placement. Compared to hierarchical device placement, GDP-one achieves an average 9.2\% speed up, and scales better to large graphs such as 8-layer NMT and 4-layer RNNLM. Importantly, with the efficient end-to-end training and sample efficient reinforcement learning algorithm, GDP-one has a 15x speed up in convergence time of the placement network over HDP.

\begin{table}[t]
	\centering
    \caption{Run time comparison between GDP-one, human expert, Tensorflow METIS, and hierarchical device placement (HDP) on six graphs (RNNLM, GNMT, Transformer-XL, Inception, AmoebaNet, and WaveNet). Graph runtime speed up is compared with Human Placement (HP) and Hierarchical Device Placement (HDP). Search speed up is the policy network training time speed up compared to HDP (reported values are averages of six runs). 
    }
    \resizebox{\textwidth}{!}{
		\begin{tabular}{c|c|c|c|c|c|c}
		\toprule
		\begin{tabular}{c} Model (\#devices)\end{tabular} & \begin{tabular}{c} \textbf{GDP-one} \\ (s) \end{tabular} &\begin{tabular}{c}  HP \\ (s) \end{tabular} & \begin{tabular}{c} METIS \\ (s) \end{tabular} & \begin{tabular}{c} HDP \\ (s) \end{tabular} & \begin{tabular}{c}Run time \\ speed up\\ over HP / HDP \end{tabular} & \begin{tabular}{c}Search \\speed up\end{tabular} \\ 
		\toprule
		2-layer RNNLM (2) & 0.234 & 0.257 & 0.355 & 0.243 & 9.8\% / 4\% & 2.95x \\
		4-layer RNNLM (4) & 0.409 & 0.48 & OOM & 0.490  & 17.4\% / 19.8\% & 1.76x \\ \hline
		2-layer GNMT (2) & 0.301 & 0.384 & OOM & 0.376 & 27.6\% / 24.9\% & 30x \\ 
		4-layer GNMT (4) & 0.409 & 0.469 & OOM & 0.520 & 14.7\% / 27.1\% & 58.8x \\ 
		8-layer GNMT (8) & 0.649 & 0.610 & OOM & 0.693 & -6\% / 6.8\% &7.35x\\ \hline
		\begin{tabular}{c} 2-layer \\ Transformer-XL (2) \end{tabular} & 0.386 & 0.473 & OOM & 0.435 & 22.5\% / 12.7\% & 40x \\
		\begin{tabular}{c} 4-layer \\ Transformer-XL (4)\end{tabular} & 0.580 & 0.641 & OOM & 0.621 & 11.4\% / 7.1\% & 26.7x\\ 
		\begin{tabular}{c} 8-layer \\ Transformer-XL (8)\end{tabular} & 0.748 & 0.813 & OOM & 0.789 & 8.9\% / 5.5\% &  16.7x \\ \hline
		Inception (2) & 0.405 & 0.418 & 0.423 & 0.417 & 3.2\% / 3\% & 13.5x \\ \hline
	    AmoebaNet (4) & 0.394 & 0.44 & 0.426 & 0.418  & 26.1\% / 6.1\% & 58.8x \\ \hline
		\begin{tabular}{c}2-stack 18-layer \\WaveNet (2)\end{tabular} & 0.317 & 0.376 & OOM & 0.354 & 18.6\% / 11.7\% & 6.67x \\ 
		\begin{tabular}{c}4-stack 36-layer \\WaveNet (4)\end{tabular} & 0.659 & 0.988 & OOM & 0.721 & 50\% / 9.4\% & 20x \\ \hline
		GEOMEAN &-&-&-&-& \textbf{16\%} / \textbf{9.2\%} & \textbf{15x} \\ \bottomrule

		\end{tabular}
	}

	\label{tab:comparison-five-tasks}
\end{table}

\subsection{Generalization}
\label{subsec:generalization}

GDP enables the training of multiple heterogeneous graphs in a single batch, sharing parameters in the graph-embedding network and the placement network. We name this training strategy \textbf{GDP-batch}. We empirically show that GDP-batch generates better placements for many workloads such as transformer-XL (7.6\%), WaveNet (15\%), and 8-layer GNMT (8\%). Table~\ref{tab:batch} compares the run time of 11 tasks using GDP-batch, with the same end-to-end architecture as described in section~\ref{subsec:ind_perf}. GDP-batch yields slightly better run time compared to GDP-one in majority of the tasks, while being only slightly worse on AmoebaNet. Compared to training graphs separately, GDP-batch reduces network parameters and enables transfer learning among different graphs. 

We further evaluate the effect of transfer learning by mixing redundant tasks in a batch. We find that mixing different graphs such as RNNLM and GNMT models with different number of layers results in both faster and better learning for RNNLM and GNMT with large number of layers (8-layer). As a matter of fact, both Placeto~\citep{Addanki2019Placeto} and HDP had problems matching human placement performance for 8-layer GNMT or 8-layer RNNLM. \textbf{With batch training, GDP is the first device placement work to match human expert performance for both 8-layer GNMT and 8-layer RNNLM. We also for the first time show that GDP-batch not only improves the search time (since we do not retrain the policy for every new graph), it can also improve the performance of the found placements.} More detailed results are shown in Appendix Table \ref{tab:batch2-5}.


\begin{table}[!htb]
    \small
	\centering
    \caption{Run time comparison on GDP-batch vs. GDP-one.}
      \centering
		\begin{tabular}{c|c|c|c}
		\toprule
		Model & Speed up & Model & Speed up \\ \midrule
	    2-layer RNNLM & 0 & Inception & 0\\ 
		4-layer RNNLM & 5\% & AmoebaNet & -5\%\\ 
		2-layer GNMT & 0 & 4-stack 36-layer WaveNet & 3.3 \% \\ 
		4-layer GNMT & 0 & 2-stack 18-layer WaveNet & 15\%\\ 
	    2-layer Transformer-XL & 7.6\% & 8-layer Transformer-XL & 1.5\%\\
	    4-layer Transformer-XL & 3\% & & \\ \bottomrule
		\end{tabular}
		\label{tab:batch}
		

\end{table}
		


\begin{figure}[!hpt]
\begin{center}
 \includegraphics[width=0.85\textwidth,trim=1cm 0cm 0.5cm 0.5cm, clip]{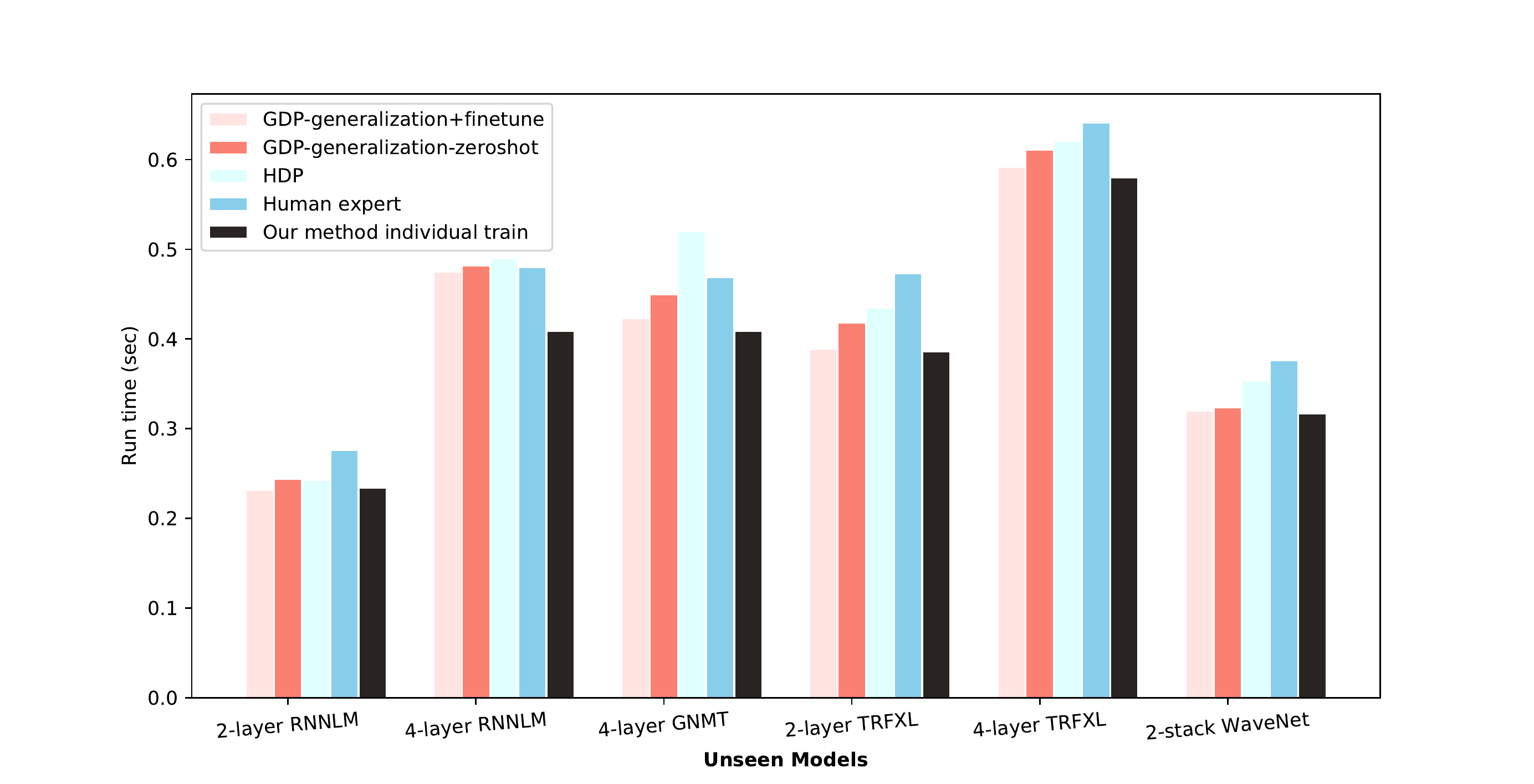}
\end{center}
\vspace{-1em}
\caption{Finetuning on hold-out graphs.}
\label{fig:unseen}
\end{figure}

\textbf{Generalization to hold-out graphs}:
Here we show another set of experiments where we treat GDP-batch as a pre-training strategy and remove the target graph from the batch training dataset. We then fine-tune the pre-trained model on the hold-out graphs for fewer than 50 steps, which takes less than one minute. We name this \textbf{GDP-generalization+finetune}. Figure~\ref{fig:unseen} shows that GDP fine-tuning for hold-out graphs outperforms human expert placement and HDP consistently on all six batch training datasets, and performs only slightly worse than GDP-one. 2-layer RNNLM and 2-stack WaveNet almost match the performance of GDP-one. We also run inference (generate placement) directly on the pre-trained model for the target hold-out graphs, and name this \textbf{GDP-generalization-zeroshot}. We find that GDP-generalization-zeroshot only marginally hurts performance as compared to GDP-generalization+finetune, while being slightly better than human placement and HDP. This indicates that both graph embedding and the learned policies transfer and generalize to the unseen data.


\textbf{Comparisons with other generalized placement approaches}: Placeto~\citep{Addanki2019Placeto}, to our knowledge, is the only other method beside GDP that shows true (and non-simulated) generalized device placement results. Direct comparison is not possible since Placeto uses a different hardware platform and different input graphs (Inception-V3, NMT, and NASNet). Placeto's search time is on average 2.65x faster than HDP, while GDP is on average 15x faster than HDP on our larger set of graphs. Apart from search time speed up, Placeto on average reduces placed graph run time by 3\% (for its different graphs and hardware) while GDP on average reduces placed graph run time by 9.2\%, compared to HDP. One advantage of GDP over Placeto is that it does not rely on any initial feasible placement. Providing a reasonable initial placement is often non-trivial for domain experts, especially for larger graphs such as 8-layer GNMT. As such, \textbf{we are the first to report superhuman results on 8-layer GNMT}. 

\subsection{Ablation Studies}
\begin{figure}[!htpb]
    \centering
    \subfloat[Comparison w/o attention.]{{\includegraphics[width=6cm,trim=1cm 0cm 1.1cm 1.32cm, clip]{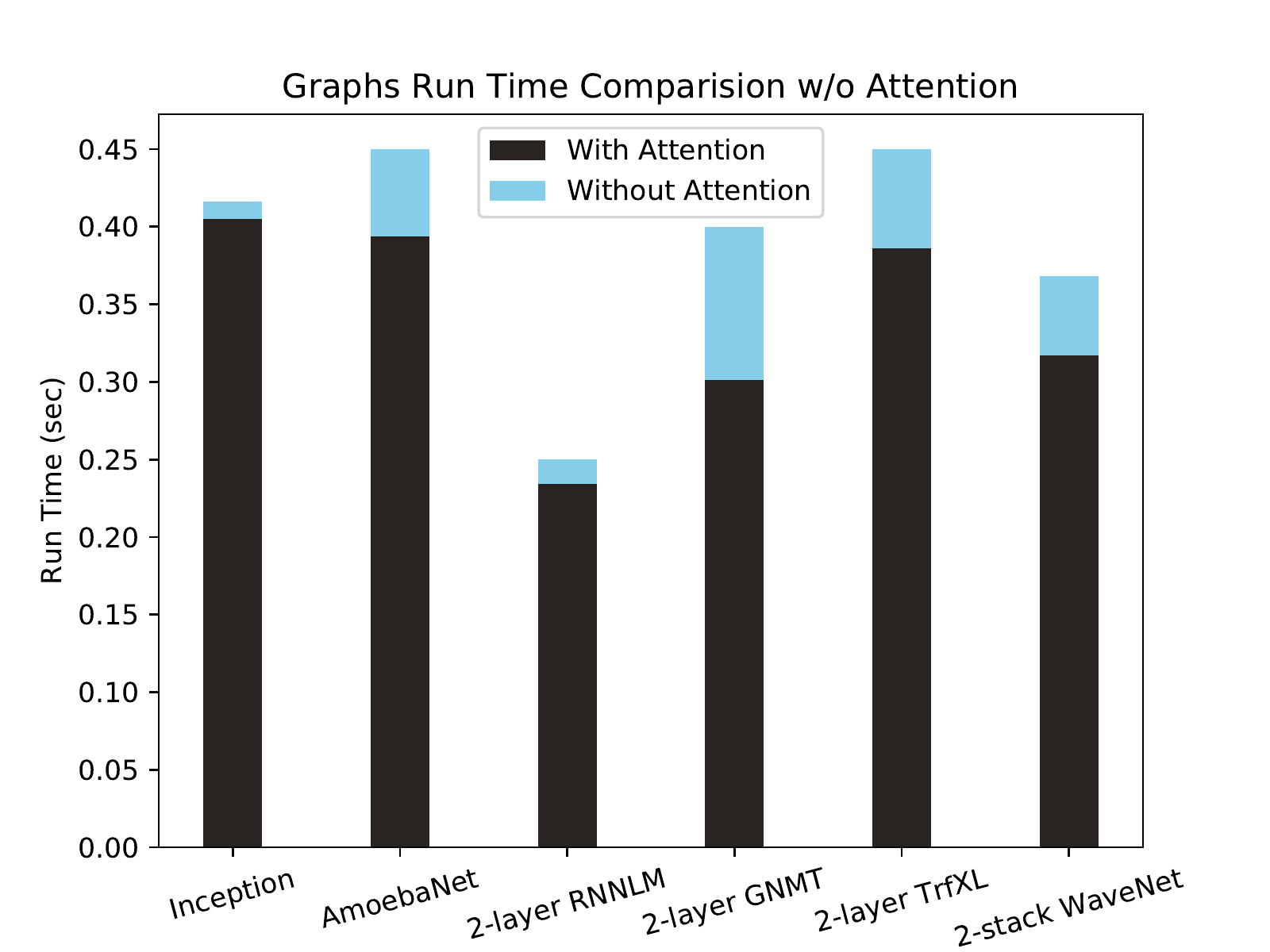} }}%
    \qquad
    \subfloat[Comparison w/o superposition.]{{\includegraphics[width=6cm,trim=1cm 0cm 1.1cm 1.32cm, clip]{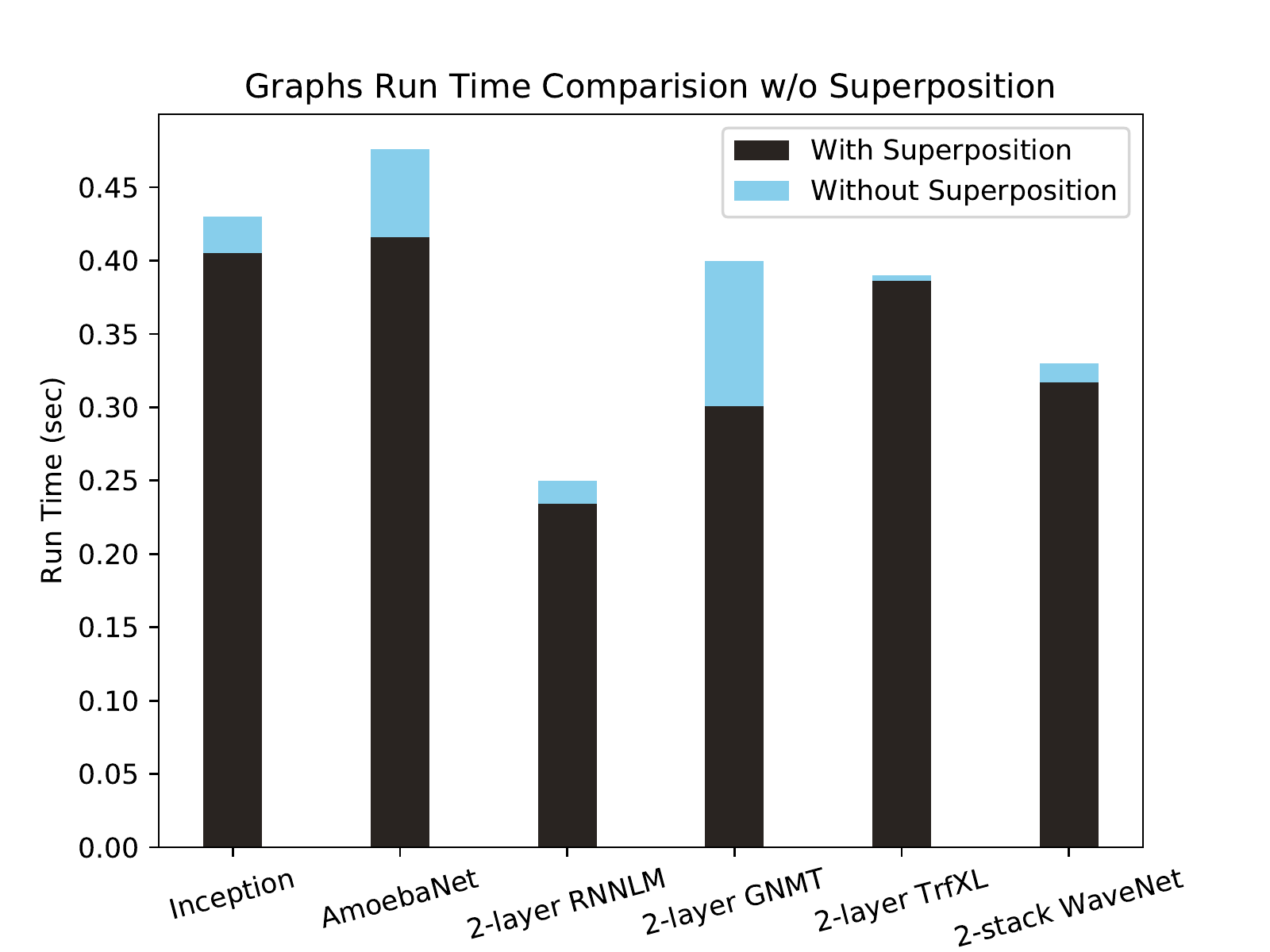} }}%
    \caption{Ablation Study on Attention and Superposition of the Placement Network.}%
    \label{fig:ablation}%
\end{figure}

\textbf{Attention and Superposition.} We did an ablation study on the attention and the superposition layer in the transformer-XL placer network. We find that attention improves placement run time by an average of 18\% compared to a placer network with no attention, and superposition improves placement run time by an average of 6.5\% where all the graphs are trained in a single batch as described in Section~\ref{subsec:generalization}. Without superposition network, batch training fails for AmoebaNet and Inception when mixing with larger RNNLM or GNMT models (4-layer). 

\begin{figure}[!t]
\begin{center}
 \includegraphics[width=0.7\textwidth,trim=0cm 0.5cm 0.2cm 0.5cm, clip]{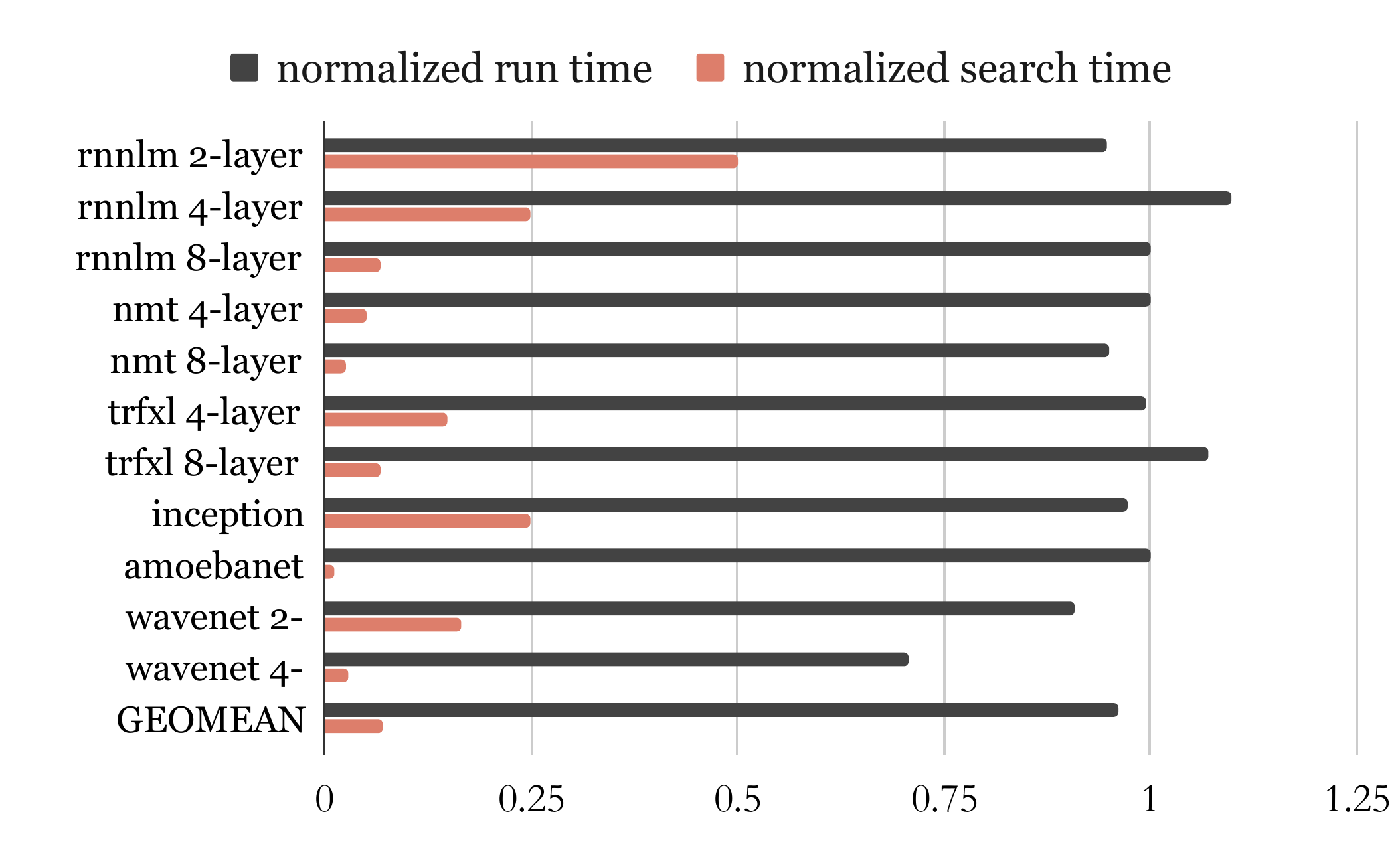}
\end{center}
\caption{Normalized run time (step time for the generated placement) and normalized training time (search time) for fine-tuning. Time is normalized to GDP without fine-tuning (training from scratch).}
\label{fig:finetune}
\end{figure}

\textbf{Pre-training graph embeddings.} We also evaluate a fine-tuning strategy by pre-training the graph embedding and placement network and fine-tuning the network on the down stream tasks. The difference here compared to Section~\ref{subsec:generalization} is that we also include the target graphs in the pre-training dataset. When GDP-batch is used as a pre-training strategy, the graph embedding and placement network assimilate meaningful graph representations and placement policies from a wide set of graphs, thus can be used as a strong baseline network for fine-tuning on downstream tasks. We compare the generated placement run time and the placement search time, normalized to GDP-one. We find that fine-tuning further reduces the the placed graph run time by an average of 5\% and placement search time by an average of 86\%, compared to GDP-one.

\section{Conclusion}
\label{conclusion}
In this paper, we present a generalized device placement strategy that uses a graph neural network and super-positioning to generalize to arbitrary and held out graphs. Through experimental evaluation over a wide set of representative graphs from different domains including computer vision, speech, and NLP, we demonstrated over 15 times faster convergence while achieving a 16\% and 9.2\% reductions in step time over human expert placement and HDP, respectively.

\subsubsection*{Acknowledgments}
TBD

\bibliography{iclr2020_conference}
\bibliographystyle{iclr2020_conference}

\section{Appendix}
\begin{table}[!htb]
	\centering
	\fontsize{6}{10}\selectfont
    \caption{Run time comparison on GDP batch training vs. the best of related methods (human expert, METIS, HDP, and GDP no batch training).}
		\begin{tabular}{|c|c|c|c|}
		\hline
	    \label{tab:batch2-5}
		Batch Setting & Model & speed up (s) \\ \hline
		 \multirow{6}{*}{Batch 2} & Inception  & 0   \\ 
		& AmoebaNet & -4.5\%  \\
		& 2-layer RNNLM & 0\\
		& 2-layer GNMT & 0 \\
		& 2-layer Transformer-XL & 6.5\%\\ 
		& 2-stack 18-layer Wavenet & 4\% \\ \hline
	    \multirow{6}{*}{Batch 3} & 2-layer RNNLM  & 0   \\ 
		& 4-layer RNNLM & 0  \\
		& 8-layer RNNLM & 4.5\%\\
		& 2-layer GNMT & 0 \\
		& 4-layer GNMT & 0 \\ 
		& 8-layer GNMT & 8\% \\ \hline
		
		Batch 4 & 3x{8-layer GNMT} & 5.1\% \\\hline
		Batch 5 & 3x{8-layer RNNLM} & 4.5\% \\\hline
		\end{tabular}
\end{table}

\end{document}